\documentclass[journal,10pt]{IEEEtran}
\usepackage{hyperref}
\usepackage{verbatim}
\usepackage{graphicx}
\usepackage{multicol}
\usepackage{amsmath}
\usepackage{cuted}
\usepackage{cite}
\usepackage{multirow}
\usepackage{hhline}
\usepackage{xcolor}
\usepackage{balance}
\usepackage{subfigure}
\usepackage{hyperref}
\usepackage{pifont}

\usepackage{stfloats}
\makeatletter

\def\hlinewd#1{%
	\noalign{\ifnum0=`}\fi\hrule \@height #1 \futurelet
	\reserved@a\@xhline}

\makeatother
	
\begin{document}
	
\title{A Full-Stage Refined Proposal Algorithm for Suppressing False Positives in Two-Stage CNN-Based Detection Methods}

\author{Qiang Guo, Rubo Zhang, Bingbing Zhang, Junjie Liu and Jianqing Liu
	
	\thanks{Qiang Guo is with College of Mechanical and Electronic Engineering, Dalian Minzu University, 116650, Dalian, China, and also with Dalian University of Technology and Postdoctoral workstation of Dalian Rijia Electronics Co., Ltd., 116630, Dalian, China. 
		(e-mail: guoqiang01486@dlnu.edu.cn).}
	\thanks{Rubo Zhang and Junjie Liu are with College of Mechanical and Electronic Engineering, Dalian Minzu University, 116650, Dalian, China. (e-mail: zhangrubo@dlnu.edu.cn, junjie1125@dlnu.edu.cn).}
	\thanks{Bingbing Zhang is with School of Computer Science and Engineering, Dalian Minzu University, 116650, Dalian, China.
		(e-mail: icyzhang@dlnu.edu.cn).}
	\thanks{Jianqing Liu is with the R\&D Department of Dalian Rijia Electronics Co., Ltd., 116630, Dalian, China. (e-mail: liujq0806@163.com).}
	\thanks{\emph{Corresponding author: Qiang Guo}.}
}

\maketitle

\begin{abstract}
	False positives in pedestrian detection remain a challenge that has yet to be effectively resolved. To address this issue, this paper proposes a Full-stage Refined Proposal (FRP) algorithm aimed at eliminating these false positives within a two-stage CNN-based pedestrian detection framework. The main innovation of this work lies in employing various pedestrian feature re-evaluation strategies to filter out low-quality pedestrian proposals during both the training and testing stages. Specifically, in the training phase, the Training mode FRP algorithm (TFRP) introduces a novel approach for validating pedestrian proposals to effectively guide the model training process, thereby constructing a model with strong capabilities for false positive suppression. During the inference phase, two innovative strategies are implemented: the Classifier-guided FRP (CFRP) algorithm integrates a pedestrian classifier into the proposal generation pipeline to yield high-quality proposals through pedestrian feature evaluation, and the Split-proposal FRP (SFRP) algorithm vertically divides all proposals, sending both the original and the sub-region proposals to the subsequent subnetwork to evaluate their confidence scores, filtering out those with lower sub-region pedestrian confidence scores. As a result, the proposed algorithm enhances the model's ability to suppress pedestrian false positives across all stages. Various experiments conducted on multiple benchmarks and the SY-Metro datasets demonstrate that the model, supported by different combinations of the FRP algorithm, can effectively eliminate false positives to varying extents. Furthermore, experiments conducted on embedded platforms underscore the algorithm's effectiveness in enhancing the comprehensive pedestrian detection capabilities of the small pedestrian detector in resource-constrained edge devices. Notably, due to the effective customization of the FRP algorithm, different combinations can be utilized based on specific task requirements to augment the model's detection capabilities to varying degrees at different computational costs, enabling it to meet the custom pedestrian detection needs of the embedded platform with diverse hardware resources.
\end{abstract}

\section{Introduction}\label{sec1}
Pedestrian detection is one of the more difficult problems in computer vision, even though image processing solutions have transitioned from traditional methods to deep learning-based methods, and the various detection methods have been constantly refreshing the gold standards on pedestrian detection benchmarks \cite{Ref1}. However, False Positives (FPs), one of the most persistent challenges in this field, where non-pedestrian objects or backgrounds are incorrectly identified as pedestrians. These issues can't be adequately solved, leading to side effects such as unnecessary braking in autonomous vehicles, inefficient resource allocation in surveillance systems, and so on \cite{Ref2}. The emergence of FPs in pedestrian detection arise from inter-class and intra-class variations, including: 1) Real-life objects are in dynamic changes, affected by variations in lighting and weather, which pose significant challenges to pedestrian detection models. They must minimize the interference of other background objects on their detection results and accurately identify human objects within complex backgrounds. 2) The diversity of pedestrian appearances, including variations in clothing, posture, and occlusion, further complicates the detection process, causing the model to struggle in distinguishing between pedestrians and backgrounds.

Recently, researchers have proposed many solutions to address this problem, but these approaches don't deeply delve into several key areas, which directly weaken the overall ability of models to suppress false positives in pedestrian detection, including:

\begin{itemize}
	\item [1)]
	Inconsistency between the training and the testing data. The proposed method mainly relies on modifying the training strategy to help the model develop the ability to effectively distinguish between positive and negative ones, and in the testing stage, the model relies on the pedestrian detection ability achieved in the training stage to accurately detect pedestrians in complex real-world scenes while effectively suppressing FP. However, the theoretical basis of this approach assumes consistency between training and testing data, which is often not realized due to the diversity and unpredictability of the real world and the limitation of finite training datasets. As a result, trained models may struggle to accurately discriminate between foregrounds and backgrounds in complex environments, resulting in FPs.
	\item [2)]
	Lack of a comprehensive solution for the model's training and testing stages. Some researchers have proposed methods to suppress pedestrian false positives specifically during the testing stage, but they can't guide the model through effective training to improve its ability to distinguish between positive and negative samples, ignoring the importance of developing strong false positive suppression capabilities during the training stage. On the other hand, certain researchers have introduced solutions focused on the model's training stage, but these solutions lack tricks for further suppressing more FPs during the inference stage. Therefore, current approaches focus primarily on enhancing the ability to suppress false positives in either the training or testing stage, but fail to provide a comprehensive solution that optimizes both stages simultaneously. Obviously, achieving a model with globally optimal pedestrian false positive suppression requires a holistic approach that ensures the model can accurately classify both positive and negative samples in both stages. 
	\item [3)]
	Computational cost issues. Current solutions lack consideration of the computational resource constraints of the target platforms, where balancing model computational cost and pedestrian false positive suppression capabilities is an issue that needs primary attention. While higher accuracy is critical for many applications, hardware resource limitations are a key constraint to deploying these approaches on platforms such as mobile systems and edge devices. Therefore, the computational feasibility of improved false positive suppression capability on resource-constrained embedded platforms is equally critical.
\end{itemize}

The above issues suggest that effectively addressing the problem of FPs in pedestrian detection needs a more comprehensive strategy for designing algorithms that work across the full stage of the model and considering the feasibility of deploying these methods under realistic conditions. Thus, this paper proposes a Full-stage Refined Proposal (FRP) algorithm to eliminate FPs in the CNN-based pedestrian detection paradigm, which can promote the model's capability to wipe out false positives during training and testing stages. In the training stage, the Training mode FRP algorithm (TFRP) utilizes a novel pedestrian proposals validated approach to guide the model training process effectively, thus constructing a model with strong false positive suppression capability. In the inference stage, two novel strategies are deployed: the Classifier-guided FRP (CFRP) algorithm integrates a pedestrian classifier into the proposal generation pipeline to produce high quality proposals through pedestrian feature evaluation; concurrently, The Split-proposal FRP (SFRP) algorithm vertically splits all proposals and sends both the original proposals and the sub-region proposals to the following subnetwork to evaluate their confidence scores, filtering out any proposals with lower sub-region pedestrian confidence scores. The CFRP and SFRP can further enhance the model's ability to wipe out pedestrian false positives in the network's inference stage to varying degrees at different computational costs. As a result, the proposed FRP algorithm helps to achieve excellent false positive suppression at all stages of the model, ensuring reliable pedestrian detection results. For embedded applications, this comprehensive approach is combined with a small pedestrian detection model: MetroNext \cite{Ref2a}, which helps the model further optimize its pedestrian detection accuracy in the training and testing stages with an acceptable increase in computational cost, helping to develop a more reliable and practical pedestrian detection system. In summary, this paper makes the following contributions:

\begin{itemize}
	\item [1)] 
	This paper analyses the reasons for false positives in the training and testing stages of the CNN-based pedestrian detection method, and proposes that the FRP algorithm suppresses the false positives in the whole stage of the pedestrian detection model to improve the model's false positive suppression ability. In the training stage, the TFRP algorithm is used to guide the model's training process to help it accurately predict the foreground/background proposal, which enhances the model's ability to recognize pedestrians. In the inference stage, the CFRP and SFRP algorithms are used in different prediction stages of the model to improve the ability to suppress false positives. According to the task requirements, different FRP algorithm combinations are adopted to construct enhanced models with better comprehensive pedestrian detection capabilities to meet different pedestrian detection task requirements.
	\item [2)]
	Various experiments are carried out on several pedestrian benchmark datasets and a metro station dataset to validate the feasibility of the proposed method. In the ablation experiments, the classical FasterRCNN detection framework is used as the baseline model, and the proposed FPR algorithm is plugged into the model for training and inference on the experimental data. The experimental results show that the FPR algorithm can effectively train the FasterRCNN to form a strong positive and negative sample classification capability in the training stage, and at the same time, it can effectively remove the FPs in the detection results in the inference stage, thus improving the accuracy of the model in the whole process stage. By combining FRP algorithms in different modes, the pedestrian detection performance of Faster R-CNN can be improved to varying degrees, with corresponding increases in computational cost to accommodate diverse pedestrian detection requirements. Experiments across  benchmark datasets and the SY-Metro dataset support that the FRP algorithm demonstrates strong generalization and cross-scene applicability, notably promoting pedestrian detection accuracy in complex real-life scenarios, even for lightweight pedestrian detection models. This makes it a better solution for pedestrian detection deployment in resource-constrained embedded scenarios.
	\item [3)]
	To validate the potential of the proposed method for application on embedded platforms. This paper combines the FRP algorithm with the small detection model: MetroNext, and conducts inference speed experiments on embedded devices. The experimental results show that the model equipped with the TFRP and SFRP algorithm obtains a better performance balance in terms of the network parameters, inference time, prediction accuracy, making it a preferred solution for embedded platforms. Combinations of the TFRP, CFRP, and SFRP algorithms help the model achieve better prediction accuracy and are suitable for solving tasks requiring high precision.
\end{itemize}

The structure of this paper is organized as follows. Section~\ref{sec2} provides an overview of the current research on suppressing false positives in pedestrian detection. Section~\ref{sec3} explains the theory of the proposed FRP algorithm, the experimental results and analysis of which are then detailed in Section~\ref{sec4}. Section~\ref{sec5} concludes the main work and contributions.

\section{Related Work}\label{sec2}
With the rapid development of big data, chip technology, and artificial intelligence, image processing-based pedestrian detection technology has become the main method to solve the challenges of pedestrian detection. The ultimate goal in this field is to reduce the pedestrian miss rate and the false positive per image (fppi), both of which are key performance metrics in pedestrian detection, and researchers have been exploiting the latest technological advances to improve these metrics steadily. This paper focuses on methods to reduce the model's false positive rate. As developed to date, they can be broadly categorized into two main directions: traditional and deep learning-based methods.

\subsection{The traditional methods}
Traditional methods dominated the field before deep learning and relied heavily on handcrafted features and machine learning techniques to tackle this challenge. Various approaches have been proposed to address this issue, each focusing on a different aspect of the problem.

One approach utilized multi-resolution infrared vision combined with matched filters to minimize false detection without using temporal or motion cues, resulting in high efficiency and low FP rates \cite{Ref3}. Another approach introduced ``chunks'' to capture pedestrian features more efficiently than conventional models, improved accuracy when cascaded with an aggregated channel feature detector, and achieved significant performance on the INRIA dataset \cite{Ref4}. In addition, Nam et al \cite{Ref5} improved the performance of orthogonal decision trees by transforming the feature representation using local correlation techniques to significantly reduce FPs at a lower computational cost, which was demonstrated on the Caltech dataset. \cite{Ref6} proposed a method using environment clustering based on false detection tendencies to create classifiers that specifically address these problems, reducing over-detection and improving the benchmark dataset's accuracy. In addition, \cite{Ref7} explored the aggregate channel characteristics of colorful thermal images and used daytime and nighttime pretrained detectors to accurately classify the photos to minimize false detections and reducing FPs.

However, traditional methods rely on hand-crafted features and are constrained by their limited representational capabilities, unable to handle complex real-world scenarios to avoid false ones effectively. As the field has evolved, these limitations have paved the way for deep learning-based approaches that can learn robust and discriminating features directly from data and more effectively incorporate contextual and complex spatio-temporal information.

\begin{figure*}[]
	\centering
	\includegraphics[width=0.6\textwidth]{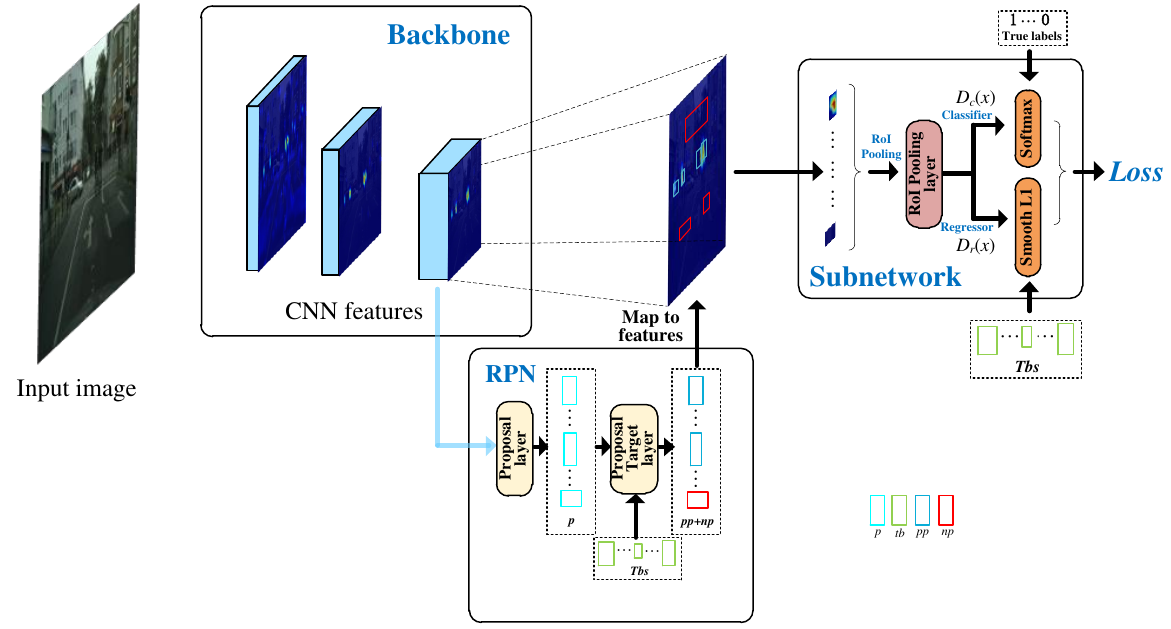}
	\caption{The detection pipeline of two-stage CNN-based pedestrian detection paradigm. Where ``p" denotes the proposals. ``tb" represents the ground truth bounding boxes. ``pp,np" represent the corresponding positive proposals and negative proposals. ``RoI Pooling" stands for the RoI Pooling layer. ``$D_c(x),D_r(x)$" denote the classifier and regressor of the subnetwork. }
	\label{FIG:1}
\end{figure*}

\subsection{The deep learning-based methods}
In pedestrian detection, the pioneering work \cite{Ref8} utilized CNN-extracted pedestrian features to train SVM classifiers to identify human objects in images. Subsequent works \cite{Ref9,Ref10, Ref11} refined this paradigm and progressively improved the detection accuracy on the benchmark datasets. To address the challenge of false positives, the researchers have designed a series of deep learning-based approaches, broadly categorized into feature enhancement, multimodal fusion, Non Maximum Suppression (NMS) strategies, and other tricks. 

\textbf{Feature enhancement}. Feature enhancement approaches strengthen and enrich pedestrian features to improve detection accuracy and reduce false positives. The Semantic Attention Fusion (SAF) mechanism enhances feature discriminability by integrating attention modules and reverse fusion blocks to generate robust semantic features and has achieved better detection results on the CityPersons dataset \cite{Ref12}. Posture Embedding Network uses human posture information to address occlusion-related false positives, combining visual and posture features to improve human-object confidence scores and achieve state-of-the-art detection results on Caltech and CityPersons datasets \cite{Ref13}. In addition, the improved RetinaNet algorithm introduces a multi-branch structure and a dual-pool attention mechanism to address multi-scale challenges, suppress irrelevant information, and improve detection accuracy on the benchmark dataset \cite{Ref14}. These approaches enhance the model's pedestrian detection capability by enriching feature representations and utilizing complementary information to suppress false positives effectively. However, while these methods greatly improve detection capabilities, their complex architectures pose challenges in terms of computational efficiency, particularly for embedded device applications. This complexity can hinder their practical application with limited computation resources in real-world scenarios.


\textbf{Multimodal fusion}. Multimodal fusion approaches reduce false positives by leveraging complementary data from multiple sensors. The IPDC-HMODL model integrates YOLOv5, RetinaNet, Kernel Extreme Learning Machine (KELM), and Hybrid Salve Swarm Optimisation (HSSO) for robust pedestrian classification, demonstrating superior performance in multimodal environments \cite{Ref15}. Full Convolutional Neural Network (FCNN) for LiDAR-camera fusion improves accuracy and precision by combining LiDAR data with multiple camera images to locate pedestrians efficiently \cite{Ref16}. TFDet employs a target-aware fusion strategy that utilizes adaptive perceptual fields and segmentation branching to enhance feature contrast for RGB-T thermal (RGB-T) pedestrian detection, and it achieves state-of-the-art results on benchmark datasets \cite{Ref17}. However, these approaches also face several challenges. Integrating multiple sensors significantly increases the development cost. In addition, adjusting the information from different sensors requires additional processing steps, thus complicating the system architecture. All these factors affect the practical deployment and scalability of pedestrian detection systems. Therefore, advanced hardware is needed to support the implementation of multimodal information fusion strategies.

\textbf{Non Maximum Suppression}. NMS is a fundamental technique in object detection, aiming at suppressing false positives by eliminating redundant bounding boxes that overlap with those of higher confidence. Recent research  has introduced several NMS variants to improve detection accuracy in crowded scenes. Representative Region NMS (R$^2$NMS) leverages less-occluded visible pedestrian parts to efficiently remove redundant bounding boxes while minimizing false positives. To identify these visible regions, a Paired-Box Model (PBM) is adopted to predict full and visible pedestrian boxes to form pairs of sample units, thereby enhancing spatial correspondence throughout the detection process \cite{Ref18}. However, this approach faces challenges in ensuring the generalization ability of the PBM to accurately predict representative areas in dynamically changing real-world scenarios. In addition, a significant amount of manual labeling effort is required to mark visible body parts during the dataset preparation. Similarly, Contextual Feature Fusion Module (CFFM) and Distance Set Non Maximum Suppression (DSNMS) utilize the spatial relationship between contextual information and overlapping frames to reduce crowd occlusion effects and suppress FPs caused by overlapping bounding boxes \cite{Ref19}. However, the above problems regarding adaptability to diverse real-world scenarios have still not been solved.

\textbf{Other tricks}. Beyond the above-mentioned methods, several novel methods have been proposed to address false positives. One of these approaches rethinks the evaluation strategy by introducing a new pipeline that simultaneously predicts the visible box of the pedestrian and uses the visible box to match the ``ignore area" to reduce misleading false positives \cite{Ref20}. Another approach, the Directed Assignment Strategy (DAS), addresses the problem of inconsistent assignment of Ground Truth (GT) bounding boxes by transforming the assignment metrics to 3D space, which optimizes the regression header and establishes a reasonable match between proposals and the GT boxes, thus reducing false ones without relying on specific thresholds \cite{Ref21}. In addition, \cite{Ref22} presents the novel model for crowded pedestrian detection, which utilizes several strategies, including distance-intersection over union loss, earth mover's distance loss, and relocation non maximum suppression is employed to select the optimal bounding boxes, thereby eliminating false positive proposal boxes. These methods improved the model's pedestrian prediction accuracy by removing FPs. However, there is a lack of exploration of the relationship between prediction accuracy gains and the resultant increase in computational cost, which limits the ability to assess the feasibility of deploying these methods in an embedded system.

In summary, while many studies have aimed to suppress FPs in pedestrian detection, to the best of our knowledge, there is still a lack of an approach suitable for deployment on embedded devices, and that achieves robust FPs suppression with less extra computational cost. To address this gap, this paper focuses on the development of a full-stage false positive suppression algorithm that aims to improve the pedestrian detection performance of two-stage CNN-based pedestrian detection methods.

\begin{figure*}[]
	\centering
	\includegraphics[width=0.6\textwidth]{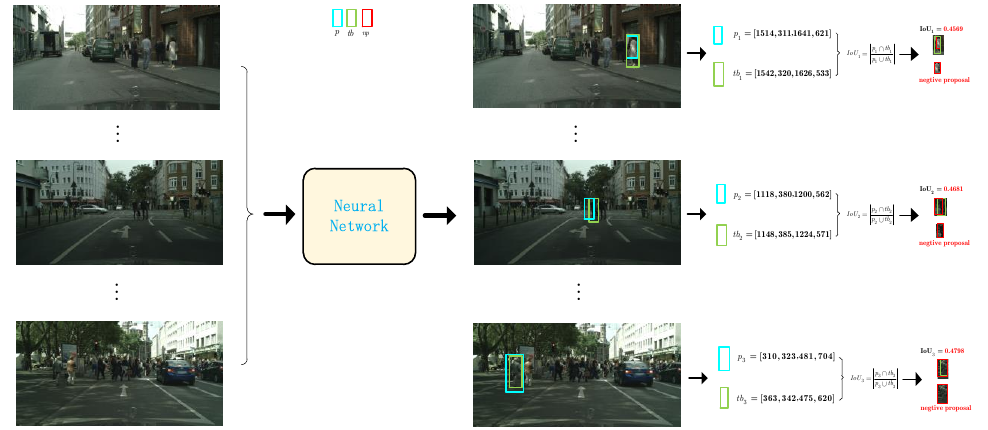}
	\caption{The misassignment samples using the IoU algorithm.}
	\label{FIG:2}
\end{figure*}

\section{Methodology}\label{sec3}
The detection pipeline of the two-stage CNN-based pedestrian detection methods is shown in Fig.~\ref{FIG:1}, where the proposal layer is used to produce a vast number of bounding boxes that cover the candidate human objects (called proposals). The following subnetwork checks whether these proposals are foregrounds and backgrounds, and the backgrounds are suppressed, and the foregrounds are treated as final detection results. In the training stage, ideally, the proposaltarget layer and grounding truths help these two network components generate appropriate foreground/background classifications, guiding the network to develop a more robust pedestrian classification capability. However, both of these components contribute to the emergence of false positives and significantly degrade the performance of the overall detection network due to the following reasons, making the FPs come from the proposal layer and subnetwork of the two-stage CNN-based detection network.

\subsection{Reasons for False Positives}
In this section, we first introduce the deficiency of the two stage CNN-based detection method in suppressing false positives from detection results. Then, the theory of the FRP algorithm is explained, and how to use it to solve this problem.

\emph{The drawback of IoU strategy.} The IoU algorithm is widely used in the network training stage for assigning positive and negative samples. However, the algorithm is also deficient, as shown in equation (1). When the denominator is much larger than the numerator, the IoU score is less than the IoU threshold, and the training sample is classified as negative ones, and this situation is common in pedestrian detection problems with multi-scale and scale variation issues. As shown in Fig.~\ref{FIG:2}, although some proposals have included pedestrian torsos, they are still misclassified as negative samples, and this misalignment weakens the classification ability of the model. During the training process, if the model can't accept accurately classified training samples, it will be difficult to develop strong pedestrian classification capabilities, making it impossible to accurately classify pedestrians during the inference stage and leading to an increase in false positive rates.

\begin{eqnarray}
	\emph{{S}}_{IoU}=\left| \frac{p\cap tb}{p\cup tb} \right|
\end{eqnarray}
where $p, tb$ denote the proposal and ground truth respectively.

\emph{The mismatch problem in the proposal selection stage.} The mismatch problem exists in the two stage CNN-based detection method, as shown in Fig.~\ref{FIG:1}, In the training stage, the proposal target layer filters out the positive and negative samples and then sends them to the following subnetwork. This means many poor-quality proposals are suppressed at this stage. However, in the inference stage, the proposaltarget layer is discarded because no ground truth bounding boxes respond to the proposals. As a result, it's unable to effectively suppress these background proposals, allowing many false ones to pass to the next detection stage, and the subnetwork faces the challenge of filtering out a lot of proposals generated in the proposal layer directly because the proposal reselection is absent. As a result, it's hard to eliminate certain complex false positive proposals during the inference stage.

\emph{Inadequate capability of subnetwork for proposal classification.} The subnetwork's primary purpose is to classify proposals into foregrounds and backgrounds accurately. However, the subnetwork struggles to remove all background proposals due to the aforementioned problems, which has weakened classification ability due to some background proposals being misclassified as foregrounds to undermines the effective training process of the model, leading to false positives in the final detection.

In short, these three factors combine to cause false positives in pedestrian detection. Therefore, a comprehensive solution is necessary to address these issues.

\subsection{The Full-stage Refined Proposal algorithm}
Aiming to address the three problems mentioned above, this paper proposes a Full-stage Refined Proposal algorithm to specifically solve these issues, which has three sub-algorithms, including: Training, Classifier-guided, and Split-proposal mode sub-algorithms, termed as TFRP, CFRP, and SFRP algorithms.

\begin{figure*}[]
	\centering
	\includegraphics[width=0.8\textwidth]{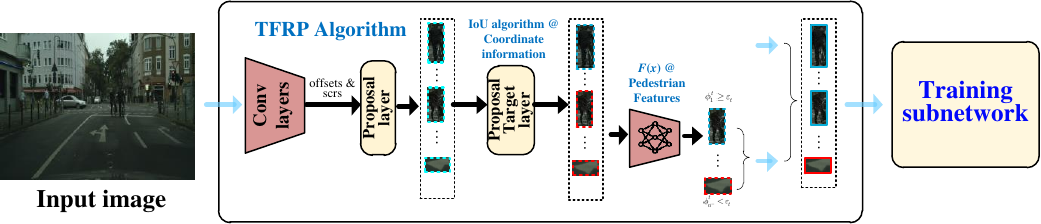}
	\caption{The processing scheme of the TFRP algorithm.}
	\label{FIG:3}
\end{figure*}

\begin{figure*}[bp]
	\centering
	\includegraphics[width=0.6\textwidth]{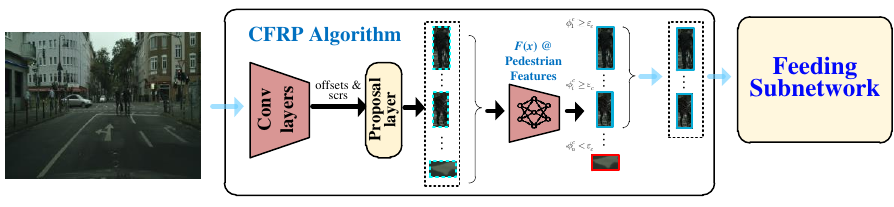}
	\caption{The processing scheme of the CFRP algorithm.}
	\label{FIG:4}
\end{figure*}
	
\subsubsection{The TFRP algorithm}
Aiming at the defects of the IoU algorithm in classifying positive and negative training samples, this paper proposes to add a human feature information evaluation step in the proposal generation stage, which combines the location information and the human feature information of the proposal to judge whether it's foreground or background. It frees from the deficiency of the IoU selection mechanism that only relies on the position information of the proposal. The evaluation of pedestrian feature information is realized by introducing a newly designed CNN-based small pedestrian classification network. The whole algorithm is introduced as follows.

In the training stage, all proposals $\emph{\textbf{P}}$ must be assigned to positive and negative samples, which are filtered out using the IoU algorithm.

\begin{eqnarray}
	\begin{aligned}
		{\emph{\textbf{P}}^{+}} &= \{p_i, IoU_i \geq \varepsilon_{IoU}, 1\leq\emph{i}\leq\emph{n}^{+}\} \\
		{\emph{\textbf{P}}^{-}} &= \{p_j, IoU_j < \varepsilon_{IoU}, 1\leq\emph{j}\leq\emph{n}^{-}\}
	\end{aligned}
\end{eqnarray}
where $\emph{\textbf{P}}^{+},\emph{\textbf{P}}^{-}$ denote the roughly selected positive and negative training proposals. $IoU_i, IoU_j$ represent  $i, j$th IoU score between their proposal and corresponding truth bounding box. $\varepsilon_{IoU}$ indicates the IoU threshold. ${n}^{+},{n}^{-}$ refers to the number of positive and negative proposals, respectively.

In the TFRP algorithm, $\emph{\textbf{P}}^{-}$ is required to filter out again to wipe out the true positive proposals, which can be achieved by introducing a small pedestrian classifier $F(x)$ to re-evaluate the pedestrian confidence score of $\emph{\textbf{P}}^{-}$.

\begin{eqnarray}
	\begin{aligned}
		\boldsymbol{\varPhi}^{t} =\{{{\varphi}^{t}_{1}},\cdots ,{{\varphi}^{t}_{i}},\cdots ,{{\varphi}^{t}_{{{n}_{-}}}}\} \\
		{{\varphi}^{t}_{i}}=F({\emph{\textbf{I}}_{p}}_{_{i}}),\ \ \ \ 1\leq i\leq {{n}_{-}}
	\end{aligned}
\end{eqnarray}
where $\emph{\textbf{I}}_{{p}_{i}}$ refers to the input image clip corresponding to $i$th proposal coordinates. 

According to the pedestrian confidence scores $\boldsymbol{\varPhi}^{t}$ of all negative proposals, $\emph{\textbf{P}}^{-}$ can be re-selected based on their pedestrian confidence scores, as shown below.

\begin{eqnarray}
	\emph{\textbf{P}}^{-}_{r}=\{{{p}^-_{i}},\ {{\phi}^{t}_{i}}<\varepsilon_{t},1\leq i\leq n_r\ \}
\end{eqnarray}

Thus, the proposal in $\emph{\textbf{P}}^{-}$ is re-evaluated by combining their pedestrian feature information and coordinate information, and the proposals rich in pedestrian feature information are removed. Finally, the new training proposals $\emph{\textbf{P}}_{tr}$ is obtained by merging $\emph{\textbf{P}}^{+}$ and $\emph{\textbf{P}}^{-}_{r}$.

\begin{eqnarray}
	{\emph{\textbf{P}}_{tr}}={\emph{\textbf{P}}^{+}}\cup \emph{\textbf{P}}_{r}^{-}
\end{eqnarray}

Therefore, the TFRP algorithm can effectively solve the problem of misassignment of training samples caused by the IoU selection strategy of assigning positive and negative ones based on the proposal coordinate information, which helps the following subnetwork to construct an accurate pedestrian classification capability by training the model on the correctly classified proposals. Fig.~\ref{FIG:3} presents the processing scheme of the TFRP algorithm.

\subsubsection{The CFRP algorithm}
In the training stage, the TFRP algorithm selects high-quality pedestrian proposals by introducing a pedestrian feature information evaluation step, which is also applicable in the testing stage to ensure the consistency of the proposal selection strategy in the training and testing stages. In the testing stage, since ground truth bounding boxes no longer exist, the proposaltarget layer will not play a role in the rough selection of proposals, and the large number of proposals generated by the proposal layer will be directly fed to the following subnetwork for processing to obtain the final detection results. To reduce the pressure faced by subnetworks to suppress these false positives in the complex field environment, this paper uses the small pedestrian classifier in the TFRP algorithm to coarsely select these proposals in its inference stage.

Firstly, the small pedestrian classifier predicts the confidence scores of all proposals $\emph{\textbf{P}}$.

\begin{eqnarray}
	\begin{aligned}
	& \boldsymbol{\varPhi}^c =\{{{\varphi}^{c}_{1}},\cdots,{{\varphi}^{c}_{i}},\cdots,{\varphi}^{c}_n\} \\ 
	& {{\varphi}^c_{i}}=F({\emph{\textbf{I}}_{p}}_{_{i}}),\ \ \ \ 1\leq i \leq n 
	\end{aligned}
\end{eqnarray}
where $n$ refers to the number of proposals.

Using the confidence scores $\boldsymbol{\varPhi}^c$, the CFRP algorithm wipes out the proposals with low pedestrian confidence scores, refining the proposals in the inference stage.

\begin{eqnarray}
	\emph{\textbf{P}}_{c}=\{{{p}_{i}},\ {{\phi}^c_{i}}\geq \varepsilon_{c},1\leq i\leq n_{c}\ \}
\end{eqnarray}

The CFRP algorithm can solves the second issue of false positives generating in the two-stage CNN-based detection methods and alleviates the mismatch problem in proposal selection stage, which can help the model to effectively remove a certain amount of false proposals during the initial proposals selection in the inference process, reducing the number of background proposals that need to be processed by the following subnetworks. Thus, it can reduce false positives rates and improve the detection accuracy of the model in the inference stage. Finally, the processing scheme of the CFRP algorithm is shown in Fig.~\ref{FIG:4}.

\begin{figure*}[bp]
	\centering
	\includegraphics[width=0.7\textwidth]{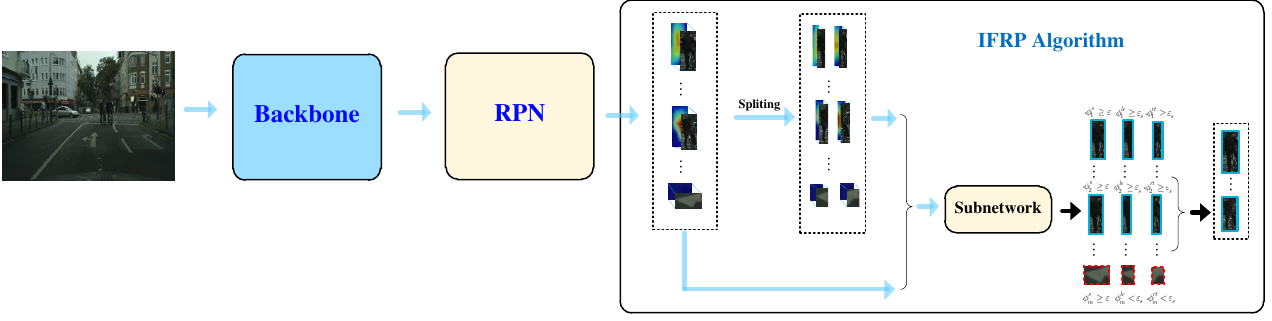}
	\caption{The processing scheme of the SFRP algorithm.}
	\label{FIG:5}
\end{figure*}

\subsubsection{The SFRP algorithm}
The TFRP and CFRP algorithms have been targeted to address the first two reasons for generating false positives in the two-stage CNN-based detection methods, and using these two algorithms, it can reduce the occurrence of false positives and improve the model's ability to accurately detect pedestrians both in the training and testing stages. These two algorithms mainly address the problem of filtering proposals in the first detection stage. The SFRP algorithm, on the other hand, addresses the issue of proposal classification of the following subnetwork in the second detection stage, further improving the model's ability to suppress false positives by enhancing its foreground/background classification capability of the subnetwork, and outputs accurate pedestrian detection results.

The subnetwork is responsible for the coordinate correction and classification of the proposals to obtain the final pedestrian detection results. The classification ability of the subnetwork directly impacts the number of false positives mixed in the detection results. Because some proposals contain some feature information similar to human objects, the subnetwork may mistakenly classify them as human objects in the final detection results. Therefore, this paper redistributes the feature information of the proposals and then processes these proposals to obtain the refined detection results.

The feature maps of proposals, which processed by the pooling operation in the RoI Pooling layer, can be represented as $\emph{\textbf{X}}_{\emph{\textbf{P}}}$, and they can be split vertically into two feature matrices by using the vertical coordinate center of proposals.

The proposals $\emph{\textbf{P}}$ can be split vertically into sub-region proposals $\emph{\textbf{P}}^{le},\emph{\textbf{P}}^{rt}$ by using their vertical coordinate center.

\begin{eqnarray}
	\begin{aligned}
		& p_i^{le}=\{{x}_{p_{i}}^{le},{y}_{p_{i}}^{le},{x}_{p_{i}}^{cen},{y}_{p_{i}}^{rt}\} \\ 
		& p_i^{rt}=\{{x}_{p_{i}}^{cen},{y}_{p_{i}}^{le},{x}_{p_{i}}^{rt},{y}_{p_{i}}^{rt}\}\\
		& {x}_{p_{i}}^{cen}={x}_{p_{i}}^{le}+({x}_{p_{i}}^{rt}-{x}_{p_{i}}^{le})/2\\
		& \emph{\textbf{P}}^{le}\cup \{p_i^{le}\}, \emph{\textbf{P}}^{rt}\cup \{p_i^{rt}\},\ \ \ \  1\leq i\leq n 
	\end{aligned}
\end{eqnarray}
%
then, a series of confidence scores for the proposals can be obtained by feeding $\emph{\textbf{X}}_{\emph{\textbf{P}}}$ and $\emph{\textbf{X}}_{\emph{\textbf{P}}^{le}},\emph{\textbf{X}}_{\emph{\textbf{P}}^{rt}}$ into the following subnetwork classification branch $D_c(x)$, respectively.

\begin{eqnarray}
	\begin{aligned}
		& {\boldsymbol{\varPhi}^{s}_{\emph{\textbf{P}}}}=\{{{\varphi}^{s}_{1}},\cdots,{{\varphi}^{s}_{i}},\cdots,{\varphi}^{s}_n\} \\ 
		& {\varphi}^{s}_{i}={{D}_{c}}(\emph{\textbf{X}}_{{{p}_{i}}}) \\ 
	\end{aligned}
\end{eqnarray}

\begin{eqnarray}
	\begin{aligned}
	& {\boldsymbol{\varPhi}^{s}_{\emph{\textbf{P}}_s}}=\{\varphi_{{{p}_{1}}}^{le},\varphi_{{{p}_{1}}}^{rt},\cdots,\varphi_{{{p}_{i}}}^{le},\varphi_{{{p}_{i}}}^{rt},\cdots,\varphi_{{{p}_{n}}}^{le},\varphi_{{{p}_{n}}}^{rt}\} \\ 
	& \varphi_{i}^{le}={{D}_{c}}(\emph{\textbf{X}}^{le}_{p_i}) \\ 
	& \varphi_{i}^{le}={{D}_{c}}(\emph{\textbf{X}}^{rt}_{p_i}) \\	
	\end{aligned}
\end{eqnarray}
where $\emph{\textbf{X}}^{le}_{P_i},\emph{\textbf{X}}^{rt}_{P_i}$ represent the left and right proposal feature maps after the splitting operation. $\emph{\textbf{P}}_s= \emph{\textbf{P}}^{le} \cup \emph{\textbf{P}}^{rt}$.

Using the pedestrian confidence scores of the proposals obtained from equations (9) and (10), the proposals are comprehensively assessed whether they are foreground or not, and when all the pedestrian confidence scores of the proposals are higher than a predetermined confidence threshold, they are recognised as foregrounds and are used as the pedestrian output detection results, otherwise these proposals are wiped out.

\begin{eqnarray}
	\emph{\textbf{P}}_{D}=\{{{p}_{i}},\ {{\phi}^s_{i}}\geq \varepsilon, \varphi_{i}^{le}\geq \varepsilon_s ,\varphi_{i}^{rt}\geq \varepsilon_s, 1\leq i\leq n_d\ \}
\end{eqnarray}

The SFRP algorithm is able to rearrange the pedestrian features of the proposals extracted by the CNN, and adopts a vertical symmetric segmentation method, which can ensure that the segmented left and right feature maps maintain the pedestrian upright characteristics, and also redistribute the original proposals features and reduce the misguidance to the classification branches of the following subnetworks when only processing the pedestrian features of the whole proposals. 

This paper adopts the vertical segmentation feature allocation strategy mainly based on the characteristics of CNN features and human object characteristics. Object classification in CNN is activated by the local key feature reinforcement of the object in the image \cite{Ref23}, when the whole feature maps are split vertically, the obtained feature maps with the pedestrian object still retain these key reinforcement feature regions, and then be sent to the classification branch of following subnetwork to still get a higher-confidence score. On the contrary, if the segmented feature maps don't contain pedestrians, they will get a lower confidence score when processed with the following subnetwork and will be regarded as background proposals to be removed. Therefore, the vertical symmetric segmentation can reduce the feature damage on the activation region considering the pedestrians characteristics. Finally, Fig.~\ref{FIG:5} presents tprocessing procedures of the SFRP algorithm


\begin{figure*}[]
	\centering
	\includegraphics[scale=.45]{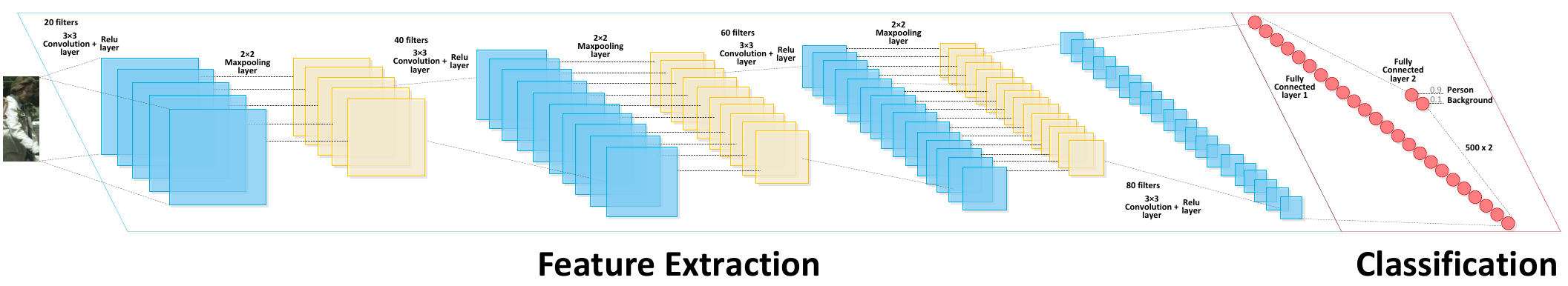}
	\caption{The architecture of the pedestrian classifier.}
	\label{FIG:6}
\end{figure*}

\subsubsection{The Small Pedestrian Classifier}
In the TFRP and CFRP algorithms, a small pedestrian classifier is required to evaluate the richness of pedestrian feature information on the proposals and remove the background proposals in which the pedestrian feature information is insufficient. Both in the training and testing stages, the small pedestrian classifier is required to have a small computational cost with better classification capability. With such a demand, this paper designs a CNN-based small pedestrian classifier.

\emph{\textbf{Component Design.}} The network mainly consists of convolutional, pooling layers, and the top fully connected layer. The convolutional layers are mainly used in CNN feature extraction. The pooling layer is used for down-sampling. The combination of pooling and convolutional layers is a simple and effective method, where the convolutional layers extract the basic features in the bottom layers and then send them to subsequent convolutional layers to form more complex and abstract features. The pooling layer reduces the spatial dimension of the feature map, which downsamples and aggregates convolutional features, making the network more robust to slight shifts in the input image. The fully connected layer maps the feature maps from the convolutional and pooling layers to the final classification result.

All convolutional layers use $3\times3$ filters and significantly reduce the number of parameters, and therefore have lower computational cost compared to other filter size choices. In addition, $3\times3$ filters can be stacked to obtain the same receptive fields as a single larger filter, while also increasing the nonlinearity of the network through the activation functions applied between layers. Thus, the use of stacked $3\times3$ convolutional filters can help the network learn complex patterns in pedestrian images with little computational burden. The layer setup of the proposed classifier is shown in Fig.~\ref{FIG:6}.

\emph{\textbf{Architecture Design.}} The input image is shaped as $64\times64$ pixels to reduces the memory footprint and the amount of computation required for convolutional operations, which is a trade-off choice after analyzing the size distribution of pedestrians in the Caltech dataset \cite{Ref43}, where most pedestrians cover 30-80 pixels, and key pedestrians attributes like body shape, head position and basic pose are still captured, so a $64\times64$ image size is sufficient for classifying pedestrians in urban life scenarios. 

The network consists of 9 layers, which are carefully determined to ensure an adequate receptive field. According to the formula (12), for a network consisting of convolutional and pooling layers, the top layer has a receptive field of more than a quarter of the $64\times64$ image. This receptive field is sufficient for the pedestrian classification problem because it can help the network capture local key features. On the other hand, a deeper network increases the computational complexity and the risk of overfitting. Therefore, a total number of 9 layers is a good choice in the overall design of the classifier.


\begin{eqnarray}
	{{r}_{n}}={{r}_{n-1}}+({f}_{n}-1)\prod\limits_{i=1}^{n-1}{{{s}_{i}}}
\end{eqnarray}
where $r_n, r_{n-1}$ represent the receptive field of $n$th layer and $(n-1)$th layer, respectively.

In the proposed classifier, residual or multi-scale feature modules are discarded. This is because these modules are not useful in shallow networks and also add extra computational effort. Thus the overall network structure of the classifier can be expressed as the following equation.

\begin{eqnarray}
	F(\emph{\textbf{X}})=(\emph{\textbf{W}}_{1}^{c}*\emph{\textbf{W}}_{1}^{p}\downarrow \emph{\textbf{W}}_{2}^{c}*\emph{\textbf{W}}_{2}^{p}\downarrow ,\cdots ,\downarrow \emph{\textbf{W}}_{9}^{fc})(\emph{\textbf{X}})
\end{eqnarray}
where $\emph{\textbf{W}}^{c},\emph{\textbf{W}}^{p}$ denote the filter weights of convolutional and pooling layers, and $\emph{\textbf{W}}^{FC}_9$ stands for the fully connected weights.

Finally, Fig.~\ref{FIG:6} clearly shows the network structure of the proposed pedestrian classifier, demonstrating the design principles and overall operation of the classification network in detail.

\begin{figure*}[]
	\centering
	\includegraphics[width=0.7\textwidth]{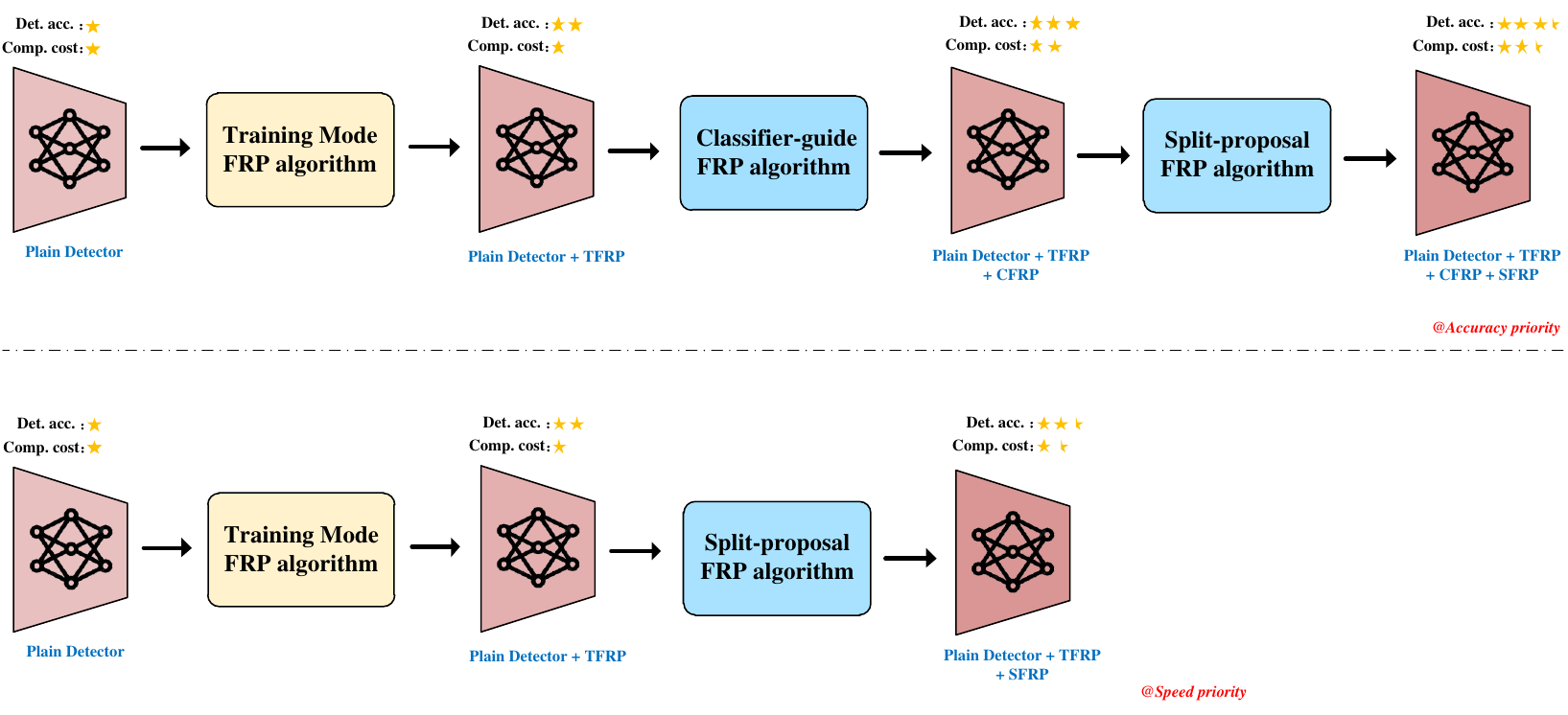}
	\caption{The schematic diagram of the detection accuracy and computational cost of FRP algorithms in different modes.}
	\label{FIG:7}
\end{figure*}

\subsection{The FRP Algorithm Application Strategy}
The FRP algorithm has three sub-algorithms, namely TFRP, CFRP, and SFRP algorithms, where TFRP guides the model to be effectively trained in the training stage and enhances the pedestrian classification ability to reduce false positives. While CFRP and SFRP algorithms reduce the number of false proposals by utilizing novel proposal re-selection strategies to redesign the proposal generation scheme in different detection stages of the model. Therefore, the TFRP algorithm does not incur extra computational cost in the model in the inference stage and belongs to the algorithmic module that is necessarily adopted in the FRP algorithm. On the other hand, CFRP and SFRP are the algorithms for selecting proposals in inference stages, which play a role in different detection stages of the model and increase the computational cost of the model to different degrees. Obviously, when the full-power FRP algorithm is selected, it will maximize the model's ability to suppress pedestrian false positives, and at the same time, the computational cost of the model increases dramatically. On the contrary, if only one mode of the FRP algorithm is employed in the inference stage, a detection performance balance between the pedestrian false positives suppression capability and the computational cost can be achieved. Therefore, it's necessary to select the appropriate combination of FRP algorithms according to the computational capability of the target platform and task requirements. Fig.~\ref{FIG:7} gives a schematic diagram of the detection accuracy and computational cost of FRP algorithms in different combination modes. 

\begin{table*}[]
	\centering
	\begin{tabular}{ll}
			\hlinewd{1pt}
			\multicolumn{2}{l}{\textbf{Algorithm 1 The detection pipeline of Full-stage Refined Proposal algorithm}} \\ \hline
			\multicolumn{2}{l}{\textbf{Algorithm 1.1 Training FRP algorithm}} \\ 
			\multicolumn{2}{l}{\begin{tabular}[c]{@{}l@{}}\textbf{Input}: $\emph{\textbf{I}},\emph{\textbf{y}}_{b}^{*},\emph{\textbf{y}}_{l}^{*}$.\\ \textbf{Output}: $\emph{\textbf{Loss}}(\emph{\textbf{I}},\emph{\textbf{y}}_{b},\emph{\textbf{y}}_{s}, \emph{\textbf{y}}_{b}^{*},\emph{\textbf{y}}_{l}^{*})$.\\ \textbf{Target}: $\underset{f:X\to {\emph{\textbf{y}}_{b}},{\emph{\textbf{y}}_{s}}}{\mathop{\min }}\,\emph{\textbf{Loss}}(\emph{\textbf{I}},\emph{\textbf{y}}_{b},\emph{\textbf{y}}_{s}, \emph{\textbf{y}}_{b}^{*},\emph{\textbf{y}}_{l}^{*})$.\\ 1. Obtain feature maps $\emph{\textbf{X}}_{\rm{backbone}}$ by using backbone network to process $\emph{\textbf{I}}$.\\ 2. Obtain proposals $\emph{\textbf{P}}$ by using RPN to process $\emph{\textbf{X}}_{\rm{backbone}}$.\\ 3. Obtain positive and negetive proposals $\emph{\textbf{P}}^{-},\emph{\textbf{P}}^{+}$ by checking their IoU scores with ground-truth boxes.\\ 4. Obtain refined proposals $\emph{\textbf{P}}^{-}_{r},\emph{\textbf{P}}_{tr}$ by using $F(x)$ to reselect $\emph{\textbf{P}}^{-},\emph{\textbf{P}}^{+}$.\\\kern 0.8pc// Initialize the refined empty proposals sets $\emph{\textbf{P}}^{-}_{r},\emph{\textbf{P}}_{tr}$ and the empty confidence set $\boldsymbol\varPhi^t$. Give training threshold $\varepsilon_{t}$\\ \kern 0.8pc4.1 \textbf{Init.} $\emph{\textbf{P}}^{-}_{r},\emph{\textbf{P}}_{tr},\boldsymbol\varPhi^t,\varepsilon_{t}$\\\kern 0.8pc// Obtain confidence scores of the negative proposals. \\ \kern 0.8pc4.2 $\boldsymbol\varPhi^t=F(\textbf{\emph{I}}_{\emph{\textbf{P}}^{-}}),\ \boldsymbol\varPhi^t=({{\varphi}^{t}_{1}},...,{{\varphi}^{t}_{n^-}})$\\\kern 0.8pc// Obtain refined negative proposals $\emph{\textbf{P}}^{-}_{r}$.\\\kern 0.8pc4.3 \textbf{for} $i\leftarrow 1$ \emph{to} $i\leftarrow n^-$ \textbf{do}\\ \kern 0.8pc4.4 \kern 0.4pc if ${{\phi}^{t}_{i}}\geq \varepsilon_t $: $\emph{\textbf{P}}^{-}_{r}\cup \{{p}^-_{i}\}$\\\kern 0.8pc// Obtain final training proposals.\\\kern 0.8pc4.5 ${\emph{\textbf{P}}_{tr}}={\emph{\textbf{P}}^{+}}\cup \emph{\textbf{P}}_{r}^{-}$\\5. Obtain training proposal feature maps $\textbf{X}_{\emph{\textbf{P}}_{tr}}$ by mapping coordinates of ${\emph{\textbf{P}}_{tr}}$ on $\emph{\textbf{X}}_{\rm{backbone}}$. \\6. Obtain $\emph{\textbf{Loss}}(\emph{\textbf{I}},\emph{\textbf{y}}_{b},\emph{\textbf{y}}_{s}, \emph{\textbf{y}}_{b}^{*},\emph{\textbf{y}}_{l}^{*})$ by using subnetwork to process $\emph{\textbf{X}}_{\emph{\textbf{P}}_{tr}}$. \end{tabular}} \\ \hline
			\multicolumn{2}{l}{\textbf{Algorithm 1.2 Testing FRP algorithm}} \\ 
			\multicolumn{2}{l}{\begin{tabular}[c]{@{}l@{}}\textbf{Input}: $\emph{\textbf{I}}$.\\ \textbf{Output}: $\emph{\textbf{y}}_{b},\emph{\textbf{y}}_{s}$.\\ \textbf{Target}: $f: \textbf{X} \to \emph{\textbf{y}}_{b},\emph{\textbf{y}}_{s}$.\\ 1. Obtain feature maps $\textbf{X}_{\rm{backbone}}$ by using backbone network to process $\emph{\textbf{I}}$\\ 2. Obtain proposals $\emph{\textbf{P}}$ by using RPN to process $\textbf{X}_{\rm{backbone}}$.\\\end{tabular}}     \\
			\begin{tabular}[c]{@{}l@{}}// The CFRP algorithm. \\ 3. Obtain new proposals $\emph{\textbf{P}}_{c}$ by suppressing proposals with low \\\kern 0.8pcconfidence scores. \\\kern 0.8pc// Initialize a empty proposals set $\emph{\textbf{P}}_{c}$ and confidence set $\boldsymbol\varPhi$,\\\kern 0.8pc give the score threshold $\varepsilon_c$.\\ \kern 0.8pc3.1 \textbf{init} $\emph{\textbf{P}}_{c}, \boldsymbol\varPhi^{c}, \varepsilon_c$\\\kern 0.8pc// Obtain confidence scores of proposals. \\ \kern 0.8pc3.2 $\boldsymbol\varPhi^{c}=F(\textbf{I}_{\emph{\textbf{P}}_c}),\ \boldsymbol\varPhi=({{\phi}^{c}_{1}},...,{{\phi}^{c}_{n}})$\\\kern 0.8pc// Obtain new proposals.\\\kern 0.8pc3.3 \textbf{for} $i\leftarrow 1$ \emph{to} $i\leftarrow n_c$ \textbf{do}\\ \kern 0.8pc3.4 \kern 0.4pc if ${{\phi}^{c}_{i}}\geq\varepsilon_c $: $\emph{\textbf{P}}_{c}\cup \{{{p}_{i}}\}$\\ 4. Obtain feature maps $\textbf{X}_{\emph{\textbf{P}}_{c}}$ by processing $\emph{\textbf{P}}_{c}$ on $\textbf{X}_{\rm{backbone}}$.\\ 5. Obtain $\emph{\textbf{y}}_{b},\emph{\textbf{y}}_{s}$ by using the subnetwork to processing $\textbf{X}_{\emph{\textbf{P}}_{c}}$.\\\\\\\\\\\\\end{tabular} & \begin{tabular}[c]{@{}l@{}} // The SFRP Algorithm.\\// Initialize empty sets $\emph{\textbf{P}}^{s}, \emph{\textbf{P}}^{le}, \emph{\textbf{P}}^{rt}$, give score thresholds $\varepsilon, \varepsilon_s$. \\3. \textbf{init} $\emph{\textbf{P}}^{s},\emph{\textbf{P}}^{le}, \emph{\textbf{P}}^{rt},  \boldsymbol{\varPhi}^{s}_{\emph{\textbf{P}}}, \varepsilon, \varepsilon_s$\\ 4. Obtain left and right proposal slices $\textbf{\emph{P}}^{s}$. \\\kern 0.8pc4.1 \textbf{for} $i\leftarrow 1$ \emph{to} $i\leftarrow n$ \textbf{do}\\\kern 0.8pc4.2 \kern 0.4pc ${x}_{p_{i}}^{cen}={x}_{p_{i}}^{le}+({x}_{p_{i}}^{rt}-{x}_{p_{i}}^{le})/2$ \\\kern 0.8pc4.3 \kern 0.4pc $p_i^{le} = ({x}_{p_{i}}^{le},{y}_{p_{i}}^{le},{x}_{p_{i}}^{cen},{y}_{p_{i}}^{rt})$ \\\kern 0.8pc4.4 \kern 0.4pc $p_i^{rt} = ({x}_{p_{i}}^{cen},{y}_{p_{i}}^{le},{x}_{p_{i}}^{rt},{y}_{p_{i}}^{rt})$\\\kern 0.8pc4.5 \kern 0.4pc $\emph{\textbf{P}}^{le}\cup \{p_i^{le}\}, \emph{\textbf{P}}^{rt}\cup \{p_i^{rt}\}$ \kern 0.4pc \\ \kern 0.8pc4.6  $\emph{\textbf{P}}^{s}=\emph{\textbf{P}}^{le} \cup \emph{\textbf{P}}^{rt}$ \\ 5. Obtain $\textbf{X}_{\emph{\textbf{P}}},\textbf{X}_{\emph{\textbf{P}}^{s}}$ by mapping coordinates of $\emph{\textbf{P}}, \textbf{\emph{P}}^{s}$ on $\textbf{X}_{\rm{backbone}}$. \\ 6. Obtain $\emph{\textbf{y}}_{b},\emph{\textbf{y}}_{s}$ by using the subnetwork to processing $\textbf{X}_{\emph{\textbf{P}}},\textbf{X}_{\emph{\textbf{P}}^{s}}$. \\ \kern 0.8pc// Obtain detection results\\ \kern 0.8pc6.1 \textbf{for} $i\leftarrow 1$ \emph{to} $i\leftarrow n$ \textbf{do}\\\kern 0.8pc6.2 \kern 0.4pc $\varphi^{s}_{i} = D_c(\emph{\textbf{X}}_{{{p}_{i}}})$; ${\varphi_{{{p}_{i}}}^{le}} = D_c(\emph{\textbf{X}}^{le}_{p_i})$; ${\varphi_{{{p}_{i}}}^{rt}} = D_c(\emph{\textbf{X}}^{rt}_{p_i})$; \\\kern 2.5pc$y_{b_i} = D_r(\emph{\textbf{X}}_{{{p}_{i}}})$\\\kern 0.8pc6.3 \kern 0.4pc if $y_{s_i}\geq\varepsilon$ \& ${\varphi_{{{p}_{i}}}^{le}}\geq\varepsilon_s$  \& ${\varphi_{{{p}_{i}}}^{rt}}\geq\varepsilon_s$: \\\kern 3pc$\emph{\textbf{y}}_{s}\cup \{\varphi^{s}_{i}\}$; $\emph{\textbf{y}}_{b}\cup \{y_{b_i}\}$\end{tabular} \\ \hlinewd{1pt}
	\end{tabular}
\end{table*}

When the plain detector is powered by FRP algorithms in different combination modes, its detection accuracy and computational cost are simultaneously increasing. When all modes of FRP algorithms are adopted to empower the detector, its detection capability and computational cost reach their maximum. It's suitable to be deployed in a hardware platform with strong computational power to accomplish the high-demand pedestrian detection task. However, for embedded platforms with weak computational capability, among the FRP algorithms in the inference stage, the CFRP algorithm has a higher computational cost compared to the SFRP algorithm, so the TFRP+SFRP algorithms can be applied to such platforms for pedestrian detection.

Finally, the pseudo-code of the FRP algorithm is listed in Algorithm 1, which gives the detailed processing steps of the FRP algorithm.


\section{Experiments and Results}\label{sec4}
This section provides details of the experimental setup and results.

\subsection{System Setup}
\emph{Experimental Platform}. Their specifications are shown in Table~\ref{Tbl1}. The selected embedded development boards are commonly used in current edge-side AI application scenarios. This allows the research work to effectively demonstrate the ability of the proposed method to accurately detect pedestrians under the constraints of existing edge-side hardware conditions, thereby validating its practical feasibility.

\setcounter{table}{0}
\begin{table}[]
	\centering
	\caption{The detailed hardware and software specifications of the experimental platforms}\label{Tbl1}
	\setlength{\tabcolsep}{0.2em}
	\begin{tabular}{lll}
		\hline
		\multirow{2}{*}{\textbf{\begin{tabular}[c]{@{}l@{}}Software \& \\ Hardware\end{tabular}}} & \multicolumn{2}{c}{\textbf{Platforms}}                           \\  \cline{2-3}
		& Workstation           & Jetson nano     \\ \hline
		CPU                                                                                       & Intel Core i9-14900KF   & ARM Cortex-A53                  \\
		MEMORY                                                                                    & 128G                   & 4G                                   \\
		GPU                                                                                       & NVIDIA RTX 4070 & NVIDIA Maxwell   \\
		Operating System                                                                          & Ubuntu 20.04     &Ubuntu 18.04\\ \hline
	\end{tabular}
\end{table}

\emph{Evaluation metrics}. The key metrics, including total parameters, inference time, miss rate, precision, and recall, are adopted to thoroughly assess the comprehensive detection performance of all detectors concerning memory usage, computational complexity, and detection accuracy. This evaluation reflects the strengths and weaknesses of these detectors in the pedestrian detection problem by comparing the experimental results across these metrics, particularly when applied to embedded platforms. Additionally, it can comprehensively evaluate the effectiveness of the FRP algorithm in enhancing the pedestrian detection capability of the plain detector and the ability of FPR algorithms with different combination modes to differentially enhance this capability, as well as their applicability under varying hardware resource constraints and task requirements.

\subsection{Datasets}
In this paper, Caltech \cite{Ref43}, CUHK\_Occ \cite{Ref44}, CityPersons \cite{Ref45}, and SY-Metro dataset are chosen as the experimental datasets. The first three datasets are widely used benchmark datasets and utilized to validate the generalization capability of the proposed methods for pedestrian detection in real-life urban cross-scenes. However, these datasets lack an important urban life scene: a metro station. To address this, a metro pedestrian dataset is collected in this paper for testing the effectiveness of the FRP algorithm in this typical scene.

\emph{SY-Metro dataset}. The dataset comprises 1503 images captured by on-board surveillance cameras deployed across various metro lines in Shenyang, China. These images have been carefully selected to comprehensively represent diverse metro scenarios under varying operational conditions. All pedestrians in these images have been labeled to support conducting pedestrian detection experiments. Therefore, the SY-Metro dataset belongs to a typical real-world subway-specific vision dataset and will be an important resource for intelligent public transportation research.

Therefore, conducting pedestrian detection experiments on these datasets with diverse pedestrian distribution characteristics makes the experimental results of the FRP algorithm more convincing in improving the performance of the plain detector for pedestrian detection. 

\subsection{Baseline detectors}
To rigorously evaluate the effectiveness of the FRP algorithm for pedestrian detection, this paper conducts comparative experiments with popular detectors including FasterRCNN \cite{Ref45a}, SSD \cite{Ref46}, Tiny YOLOV3 \cite{Ref47}, FPN \cite{Ref48}, Pelee \cite{Ref49}, EfficientDet \cite{Ref50} and Swin Transformer \cite{Ref51}. These state-of-the-art models span different architectural paradigms and ensure an objective and effective comparative evaluation of FRP for improving the detection capability of the plain model with different combination modes. Based on the results of the comparative experiments, it's possible to verify the different enhancement capabilities of the proposed method for pedestrian detection in different combination modes, and to construct pedestrian detection networks with different performance strengths using the FRP algorithm. This comparative analysis can highlight the advantages and limitations of FRP algorithms in different modes and provide experimental results to support their practicality in real application scenarios.

\begin{table}[!h]
	\centering
	\caption{Results of Citypersons dataset from FasterRCNN using the FRP algorithm.}\label{Tbl2}
	\setlength{\tabcolsep}{0.2em}
	\scalebox{0.9}{\begin{tabular}{cccccccc}
			\hline
			\textbf{\begin{tabular}[c]{@{}c@{}}Model @ \\ CityPersons\end{tabular}} & Backbone               & TFRP          & CFRP & SFRP &MR(\%)       & Params(M)    & GPU(ms/frame) \\ \hline
			\multirow{2}{*}{FasterRCNN}                                             & \multirow{2}{*}{VGG16} &  \ding{51}    &  \ding{55}    &  \ding{51}    & 73.63(-1.41) & 137.1(+0)    & 57(+4)       \\  
			&                        &  \ding{51}    &  \ding{51}    &  \ding{51}    & 73.63(-1.87) & 137.1(+0.71) & 57(+30)       \\ \hline
	\end{tabular}}
\end{table}
\subsection{The ablation experiment}
In this paper, the widely used detector FasterRCNN and the popular pedestrian detection benchmark dataset CityPersons are employed in the ablation experiments to verify the feasibility of the FRP algorithm in solving pedestrian detection problems. In particular, the improvement of the detection ability and computational cost of the plain detector by using the FRP algorithm under different algorithm combination modes is explored. The detailed experimental results are shown in Table~\ref{Tbl2}, where ``MR" denotes log average miss rate (the lower the better), ``Params" denotes the total parameters of the network (``M" means millions), and ``GPU" denotes the time spent per frame in milliseconds (ms/frame). Values in parentheses indicate the change in the corresponding performance metrics compared to the plain detector without the help of the FRP algorithm.

\begin{table*}[]
	\centering
	\caption{Experimental results of the FRP-enhanced MetroNext and baseline detectors on the benchmark datasets}\label{Tbl3}
	\setlength{\tabcolsep}{0.2em}
	\begin{tabular}{lllllllllll}
		\hline
		\multicolumn{4}{l}{\multirow{2}{*}{\textbf{Models}}}                & \multirow{2}{*}{\begin{tabular}[c]{@{}l@{}}Params\\ (M)\end{tabular}} & \multicolumn{2}{l}{\textbf{CUHK-Occ}}                                                                        & \multicolumn{2}{l}{\textbf{Caltech}}                                                                         & \multicolumn{2}{l}{\textbf{Citypersons}}                                                                     \\ \cline{6-11} 
		\multicolumn{4}{l}{}                                                &                                                                       & \begin{tabular}[c]{@{}l@{}}MR\\ (\%)\end{tabular} & \begin{tabular}[c]{@{}l@{}}GPU\\ (ms/frame)\end{tabular} & \begin{tabular}[c]{@{}l@{}}MR\\ (\%)\end{tabular} & \begin{tabular}[c]{@{}l@{}}GPU\\ (ms/frame)\end{tabular} & \begin{tabular}[c]{@{}l@{}}MR\\ (\%)\end{tabular} & \begin{tabular}[c]{@{}l@{}}GPU\\ (ms/frame)\end{tabular} \\ \hline
		\multicolumn{4}{l}{Swin Transformer}                                & 45.31                                                                 & 24.80                                             & 65                                                      & 58.72                                             & 63                                                      & 60.22                                             & 63                                                    \\
		\multicolumn{4}{l}{FPN}                                             & 42.12                                                                 & 28.01                                             & 83                                                      & 59.68                                             & 83                                                     & 64.28                                             & 85                                                      \\
		\multicolumn{4}{l}{FRCNN VGG16}                                     & 136.69                                                                & 36.59                                             & 57                                                       & 64.77                                             & 56                                                       & 73.63                                             & 57                                                       \\
		\multicolumn{4}{l}{SSD512}                                          & 23.75                                                                 & 35.81                                             & 34                                                       & 69.26                                             & 35                                                       & 79.41                                             & 35                                                       \\
		\multicolumn{4}{l}{Tiny YOLOV3}                                     & 8.66                                                                  & 53.33                                             & 2                                                        & 77.63                                             & 2                                                    & 86.82                                             & 2                                                        \\
		\multicolumn{4}{l}{Pelee}                                           & 5.29                                                                  & 41.28                                             & 13                                                       & 74.42                                             & 13                                                       & 83.96                                             & 13                                                       \\
		\multicolumn{4}{l}{EfficientDet}                                    & 3.90                                                                  & 40.62                                             & 32                                                       & 71.49                                             & 32                                                       & 80.49                                             & 33                                                       \\\hline
		\multicolumn{1}{c}{\multirow{4}{*}{MetroNext}} & TFRP & CFRP & SFRP & \multicolumn{7}{l}{}                                                                                                                                                                                                                                                                                                                                                                                               \\
		\multicolumn{1}{c}{}                           &  \ding{55}    &  \ding{55}    &  \ding{55}    & 4.56                                                                  & 37.81                                             & 19                                                       & 64.82                                             & 17                                                       & 71.87                                             & 25                                                       \\
		\multicolumn{1}{c}{}                           &   \ding{51}   &   \ding{55}   &   \ding{51}   & +0                                                                  & -1.15                                             & +4                                                       & -0.38                                             & +4                                                       & -2.60                                             & +5                                                       \\
		\multicolumn{1}{c}{}                           &  \ding{51}    &  \ding{51}    &  \ding{51}    & +0.71                                                                  & -1.68                                             & +18                                                       & -0.61                                             & +21                                                       & -3.11                                             & +23                                                       \\ \hline
	\end{tabular}
\end{table*}

In this paper, two combination modes of FRP algorithms are adopted to equip FasterRCNN for hunting pedestrians, namely the TFRP+SFRP (Compact mode FRP) algorithm and the TFRP+CFRP+SFRP (Full mode) algorithm. First of all, the TFRP algorithm is the first-adopted algorithm in the FRP algorithm because it's used to guide the network to learn effectively in the training stage to form an accurate ability to distinguish pedestrians from the background, and improving the pedestrian detection ability of the model in the inference stage without extra computational burdens. Secondly, among the different modes of FRP algorithms, the SFRP algorithm relies on the subnetwork to achieve false proposals suppression by processing the feature rearranged proposals, and its algorithmic computational complexity is lower compared to the CFRP algorithm, which introduces a cascaded pedestrian classification network in the algorithm processing procedures. Therefore, this paper combines the TFRP algorithm with the SFRP algorithm to establish the TFRP+SFRP algorithm, which has a certain ability to improve the detection capability, but with lower algorithmic complexity, so as to meet the application requirements of low computing power platforms and simple pedestrian detection tasks. For vision tasks with high computational power and strict pedestrian detection requirements, the TFRP+CFRP+SFRP algorithm, which includes all modes of the FRP algorithm, is established to meet the requirements of these detection tasks. 

As can be seen from the experimental results in Table~\ref{Tbl2}, the MR result of FasterRCNN is 73.63\% for a total parameter of 137.1M and a GPU inference speed of 57ms/frame. With the help of TFRP+SFRP algorithms, the MR result decreases by 1.41\%, while the detection speed increases by only 4ms, which indicates that the compact mode FRP algorithm effectively improves the detection accuracy of the plain detector with little extra computational burden. On this basis, when the CFRP algorithm is introduced again in the model inference stage, the MR result can be further reduced by 1.87\%. Due to the incorporation of the cascaded pedestrian classification network, the total number of model parameters and the inference time increase by 0.71M and 30ms, respectively, which suggests that the full-mode FRP algorithm needs to add some extra computational cost when effectively reducing the model's MR result. Therefore, the experimental results show that the compact-mode FRP and full-mode FRP algorithms can effectively improve the detection accuracy of the model with additional computational cost, and the full-mode FRP algorithm improves the detection accuracy more significantly. 

\begin{figure}[!h]
	\centering
	\includegraphics[width=0.35\textwidth]{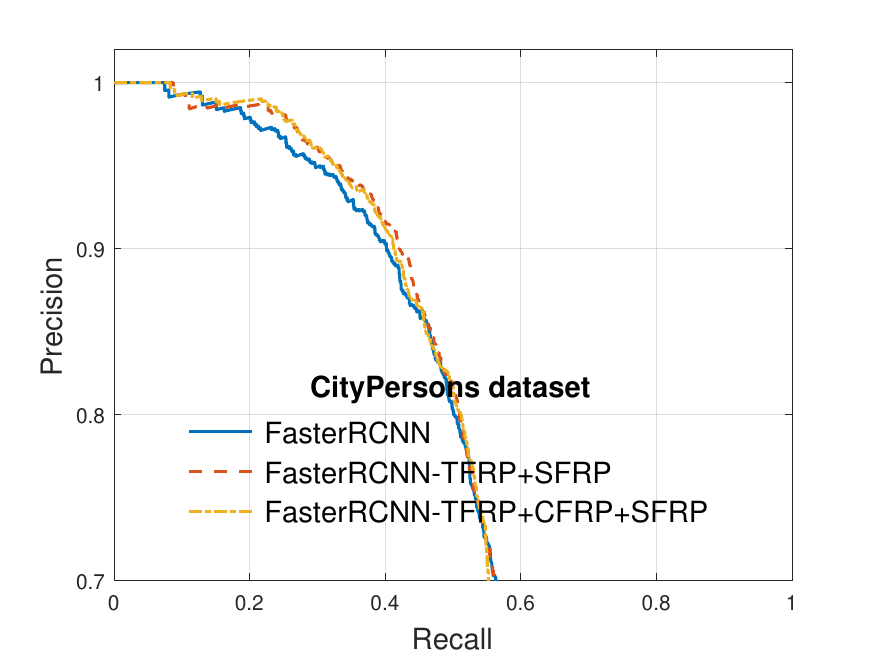}
	\caption{Precision-recall curves of the FRP-enhanced FasterRCNN and baseline detectors on the CityPersons datasets.}
	\label{FIG:8}
\end{figure}

The corresponding precision recall curve of the FasterRCNN on the ablation experiment has been drawn in Figure~\ref{FIG:8}, which demonstrates the effect of the two-mode FRP algorithm to promote the detection accuracy of the plain detector.

\subsection{Benchmark experiments}
The proposed FRP algorithm is evaluated on benchmark datasets including CUHK-Occ, Caltech and CityPersons, and images in these datasets reflect real-world urban scenes with various lighting conditions, pedestrian density, and background clutter, which provides a comprehensive evaluation of the FRP algorithm's ability to detect pedestrians in real-world scenarios, and verifies whether the algorithm is highly generalization for improving pedestrian detection capability in complex urban scenes. The detailed experiment results are presented in Table~\ref{Tbl3}, as can be seen that:

\begin{itemize}
	\item [1)]
	In the three benchmark datasets, the plain MetroNext has achieved better detection results, the MR results of which are 37.81\%, 64.82\%, 71.87\%. With the help of the compact-mode FRP and full-mode FRP algorithms, their MR reductions reach up to 2.6\% and 3.11\% (the lowest values are 0.38\% and 0.61\%, respectively). That is, the MR values are reduced to different degrees on each dataset, indicating that the FRP algorithms have the ability to improve the detection accuracy of the plain model in complex multi-scenarios. Meanwhile, the two FRP algorithms also improve the computational cost of the plain model to different degrees, with the maximum increase in inference time and total parameters are 23ms and 0.71M, respectively. Although this introduces some additional model's computation costs, they are acceptable considering the significant improvement in detection capabilities.
	\item [2)]
    In the citypersons and cuhk-occ datasets, the two FPR algorithms both effectively reduce the MR values beyond their performance on the Caltech dataset. The reason for this phenomenon is that these two datasets contain more pedestrian crowding situations compared to the Caltech dataset. When detecting crowded pedestrians, the IoU threshold has a direct impact on the final detection results. However, the TFRP and CFRP algorithms eliminate the drawback of relying only on the IoU threshold to select the proposals by introducing the pedestrian feature evaluation procedure into the proposal generation, thus decoupling the IoU algorithm from the proposal generation process. This helps the plain detector generate more high-quality pedestrian proposals, effectively enhancing the model's pedestrian detection accuracy, which demonstrates that the FRP algorithms are more effective in addressing the problem of occluded pedestrian detection.
    \item [3)]
    This paper combines the two FRP algorithms with MetroNext, respectively. Combining the compact-mode FRP algorithm with MetroNext, a pedestrian detector with balanced detection performance is constructed with an MR result of 69.27\% ( Citypersons dataset), an inference speed of 30ms, and total parameters of 4.56M. When the full-mode FRP algorithm is combined with MetroNext, an accuracy-oriented pedestrian detector is constructed, whose MR value is reduced to 68.76\%, the inference speed is 48ms, and the total number of network parameters is 5.27M. Therefore, a comprehensive evaluation considering inference speed, model size, and detection accuracy, it can be seen that the FRP-enhanced MetroNext demonstrates a more competitive overall pedestrian detection performance compared to the other baseline detectors. For example, although Swin Transformer has excellent detection accuracy, its large model size and high computational complexity restrict the deployment of the model in practical applications. In addition, the FRP algorithm demonstrates wide-ranging applicability across various model scales, and even for a small model like MetroNext, which can effectively improve the model's detection capability with an acceptable increase in computational cost, further highlighting the strengths and practical merits of the FRP algorithm.

\end{itemize}


\begin{figure}[!h]
	\centering
	\subfigure[CUHK-Occ dataset]{
		\includegraphics[scale=.35]{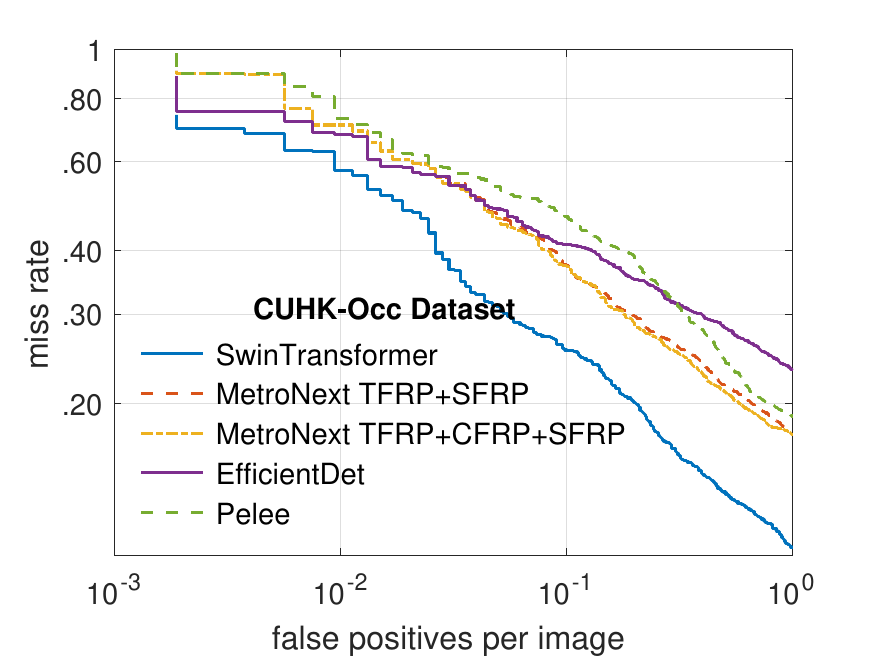}
	}
	\subfigure[Caltech dataset]{
		\includegraphics[scale=.35]{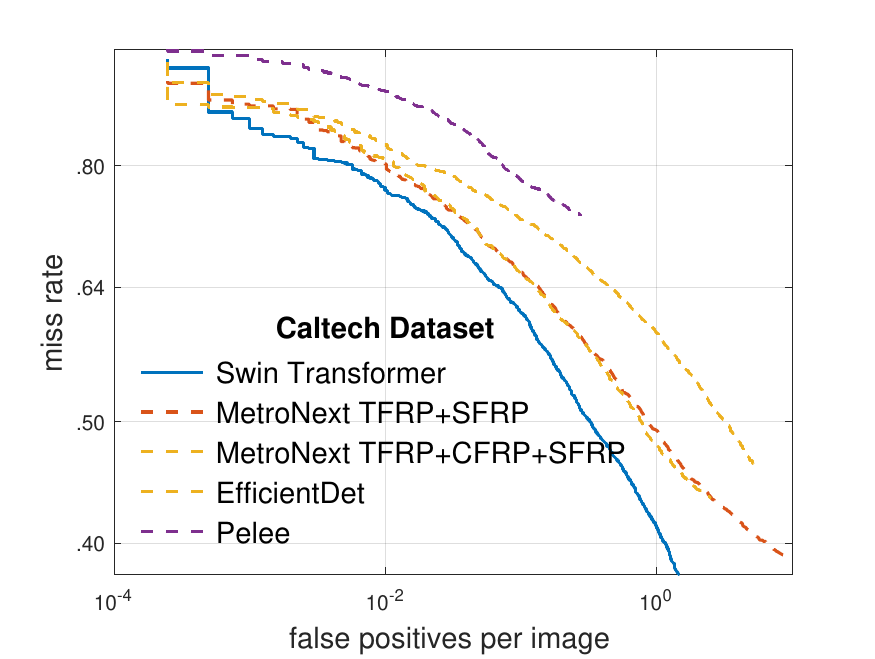}
	}
	\subfigure[CityPersons dataset]{
		\includegraphics[scale=.35]{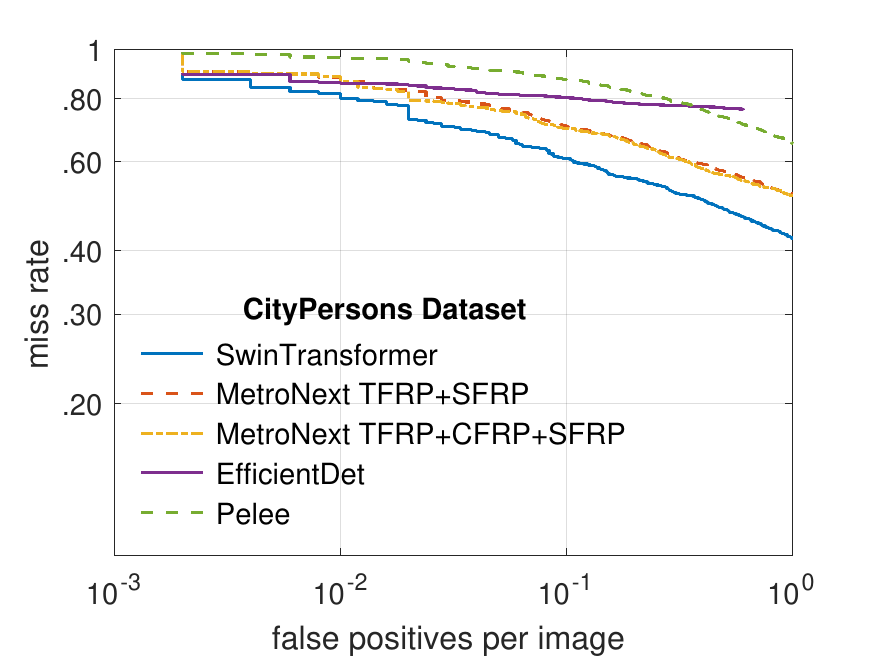}
	}
	\caption{MR curves of the FRP-enhanced MetroNext and baseline detectors on benchmark datasets.}
	\label{FIG:9}
\end{figure}

\textbf{Discussion} All detectors are trained and evaluated using a consistent dataset partition, which ensures that the performance evaluation of each detector is fair and comparable. The proposed FRP algorithm has demonstrated detection performance gains on various benchmark datasets, demonstrating the robustness of the FRP algorithm in promoting the model's capability in classifying pedestrians and backgrounds. By integrating the FRP algorithm, the FRP-enhanced MetroNext achieves superior efficiency metrics, including a reduced model size (5.27M total parameters) and faster inference speed (21ms/frame), but there is still a gap in detection accuracy compared to larger-scale baseline models. This accuracy-performance trade-off stems from the inherent limitations of the FRP algorithm's compact designing strategies, which restrict its ability to retain the full range of semantic features necessary for accurate pedestrian/background classification. Nevertheless, the FRP algorithm offers a competitive solution for real-world pedestrian detection tasks, and the compact-mode FRP algorithm is particularly well suited for deployment on resource-constrained edge devices, where low latency outweighs minor sacrifices in accuracy.


The miss rate curves of the FRP-enhanced MetroNext, Swin Transformer, EfficientDet, and Pelee are plotted in Figure~\ref{FIG:9}, where the vertical axis represents the miss rate, with a range interval of [0, 0.1], and the horizontal axis represents the false positive per image values (fppi), spanning the interval of [$10^{-3}$, $10^{0}$]. The curves reflect the changes in the detector's miss rate and fppi with the variation of the threshold value. The lower the curve is, the better the pedestrian detection capability. The legend in the figure indicates the log average miss rate of these model. 

\begin{figure}[!h]
	\centering
	\subfigure[CUHK-Occ dataset]{
		\includegraphics[scale=.35]{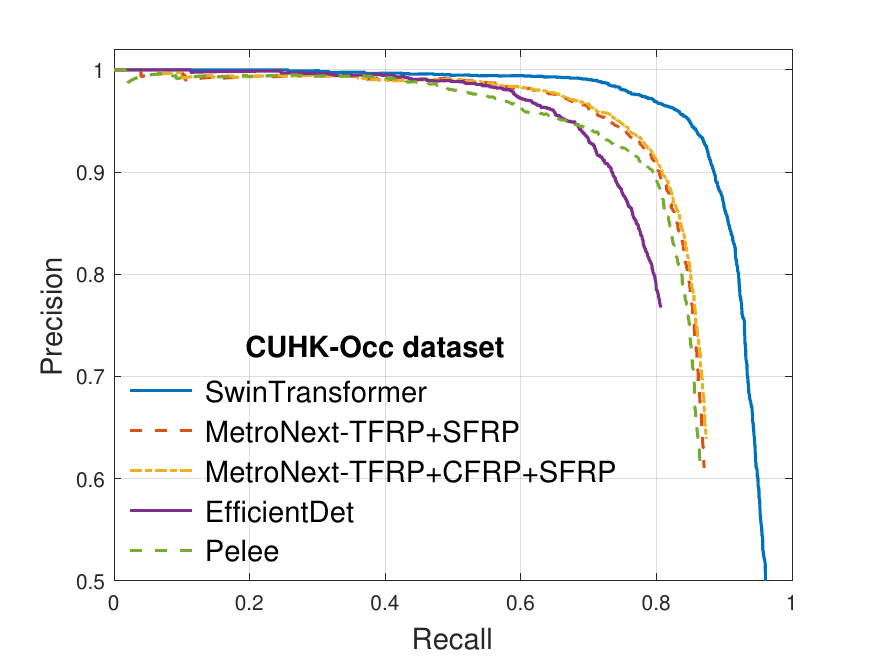}
	}
	\subfigure[Caltech dataset]{
		\includegraphics[scale=.35]{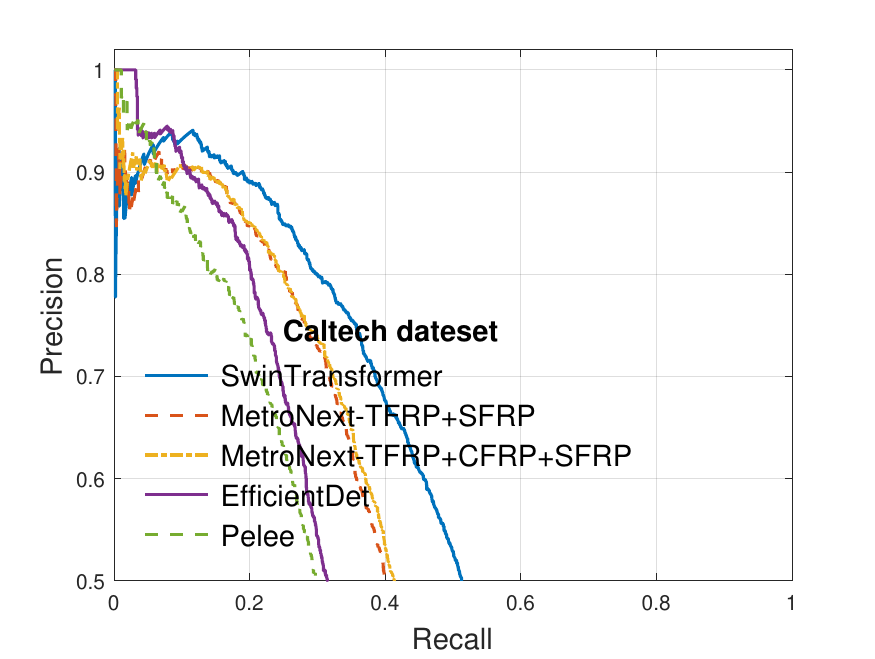}
	}
	\subfigure[CityPersons dataset]{
		\includegraphics[scale=.35]{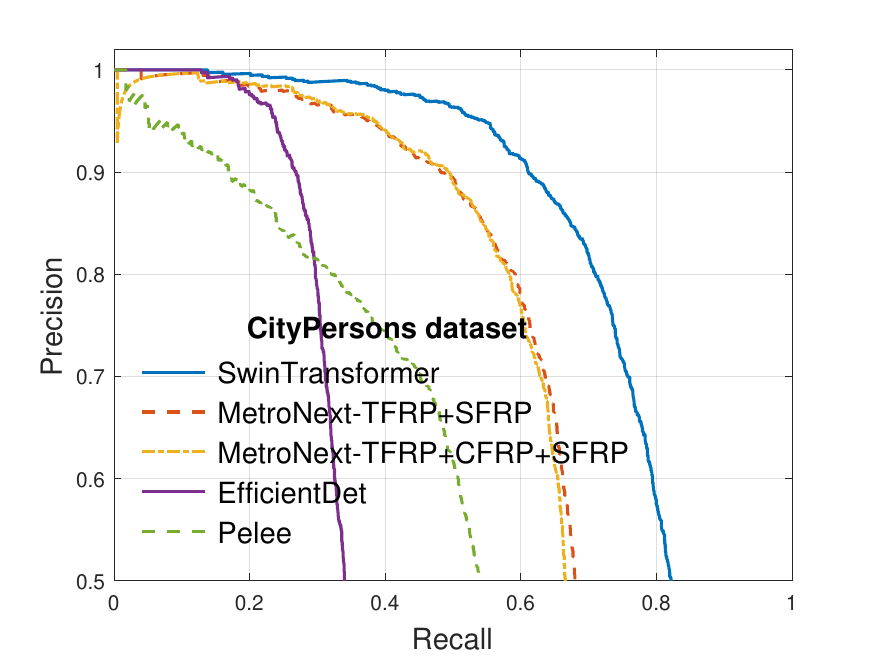}
	}
	\caption{Precision-recall curves of the FRP-enhanced MetroNext and baseline detectors on the benchmark datasets.}
	\label{FIG:10}
\end{figure}

The precision recall curves for these detectors are presented in Figure~\ref{FIG:10}, with precision and recall axes spanning the intervals $[0.5,1.0]$ and $[0,1.0]$, respectively. These curves illustrate the pairs of precision and recall as the threshold varies, serving to quantify each detector's capability to accurately identify true positives while suppressing false ones. A higher-positioned curve implies better detection performance.

As illustrated in Figures~\ref{FIG:9} and~\ref{FIG:10}, the curves of the FRP-enhanced MetroNext are in a dominant position. This indicates that the FRP-enhanced MetroNext achieves better pedestrian detection performance, and the proposed FRP algorithm helps the plain MetroNext to improve the localization precision and reduce the false positive rates. Considering the relatively small increase in computational cost brought by the FRP algorithm, it supports the feasibility of utilizing the FRP algorithm to build a practical pedestrian detector to provide a reliable pedestrian detection technology solution in real-life scenarios, where reliable pedestrian detection must balance accuracy with computational pragmatism.

\begin{table}[!h]
	\centering
	\caption{Experimental results of the FRP-enhanced MetroNext and baseline detectors on SY-Metro datasets}\label{Tbl4}
	\setlength{\tabcolsep}{0.2em}
\begin{tabular}{lllllll}
	\hline
	\multicolumn{4}{l}{\multirow{2}{*}{\textbf{Models}}} & \multirow{2}{*}{\begin{tabular}[c]{@{}l@{}}Params \\ (M)\end{tabular}} & \multicolumn{2}{l}{SY-Metro}                                                                                   \\ \cline{6-7} 
	\multicolumn{4}{l}{}                                 &                                                                        & \begin{tabular}[c]{@{}l@{}}MR\\(\%)\end{tabular} & \begin{tabular}[c]{@{}l@{}}GPU \\ (ms/frame)  \end{tabular} \\ \hline
	\multicolumn{4}{l}{Swin Transformer}                 & 45.31                                                                  & 65                                                       & 18.11                                              \\
	\multicolumn{4}{l}{FPN}                              & 42.12                                                                  & 83                                                       & 21.33                                              \\
	\multicolumn{4}{l}{FRCNN VGG16}                      & 136.69                                                                 & 57                                                        & 29.99                                              \\
	\multicolumn{4}{l}{SSD512}                           & 23.75                                                                  & 34                                                        & 26.16                                              \\
	\multicolumn{4}{l}{Tiny YOLOV3}                      & 8.66                                                                   & 2                                                         & 42.29                                              \\
	\multicolumn{4}{l}{Pelee}                            & 5.29                                                                   & 13                                                        & 32.25                                              \\
	\multicolumn{4}{l}{EfficientDet}                     & 3.90                                                                   & 32                                                        & 27.27                                              \\ \hline
	\multirow{4}{*}{MetroNext}   & TFRP  & CFRP  & SFRP  & \multicolumn{3}{l}{}                                                                                                                                                                    \\
	 &  \ding{55}    &  \ding{55}    &  \ding{55}    & 4.56                                                                   & 26.70                                                        & 18                                              \\
	&   \ding{51}   &   \ding{55}   &   \ding{51}   & +0                                                                   & -2.26                                                        &+3                                               \\
	&  \ding{51}    &  \ding{51}    &  \ding{51}    & +0.71                                                                   &-2.32                                                        &+27                                              \\ \hline
\end{tabular}
\end{table}

\subsection{The SY-Metro experiment}
To further test the generalization ability of the FRP algorithm in improving pedestrian detection across more diverse real-world scenarios, this paper conducts experiments in a metro station dataset, SY-Metro, which is rarely involved in the aforementioned benchmark datasets. The detailed experimental results are shown in Table~\ref{Tbl4}, as showing:

\begin{itemize}
	\item [1)]
	The experimental results on the SY-Metro dataset reveal that the two FRP algorithms promote the pedestrian detection capability of the plain MetroNext to varying degrees. Specifically, the compact-mode FRP algorithm reduces the MR by 2.26\%, and the full-mode FRP algorithm cuts down the MR value by 2.32\%. This further narrows the gap with other large-scale models in terms of MR values, while requiring fewer computational resources, indicating that FRP-enhanced MetroNext is a competitive pedestrian detection model with better overall detection performance.
	\item [2)]
	Compared to the complicated outdoor scenes, the metro station represents a more settled scene. As a result, all detectors achieve lower MR values in this scene, showing better pedestrian detection accuracy. At the same time, this scene is also a highly crowded place, so the FRP algorithm also shows a better ability to improve the detection performance of plain models, showing a decrease of about 2\% in the MR value. This performance gain highlights that the FRP algorithm has robust generalization ability to improve the model's pedestrian detection performance across different crowded scenes.
\end{itemize}

\begin{figure}[!h]
	\centering
	\includegraphics[scale=.35]{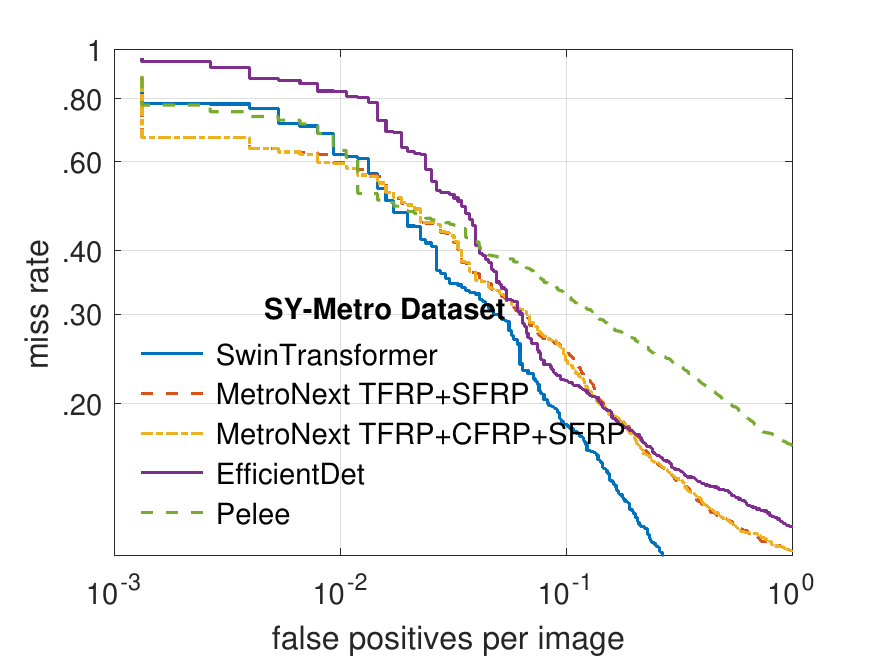}
	\caption{the FRP-enhanced MetroNext and baseline detectors on SY-Metro dataset}
	\label{FIG:11}
\end{figure}

Figure~\ref{FIG:11} visualizes the miss rate versus fppi curves for the FRP-enhanced MetroNext and baseline detectors on the SY-Metro dataset. As is shown in Fig~\ref{FIG:11}, this graph clearly illustrates the detection capabilities of each model in recognizing metro passengers. Compared with other competitors, our model attains a better detection performance in metro passenger detection, positioning it as a competitive solution for passenger detection in metro stations.

\subsection{The experiment on the embedded development board}
The image processing efficiency of the FRP algorithm is a key performance metric for addressing the pedestrian detection problem in edge devices. In this paper, the Jetson Nano is utilized as the experimental platform to empirically measure the inference speed of the FRP algorithm under different modes. The detailed results are clearly presented in Table~\ref{Tbl5}, showing that:

\begin{itemize}
	\item [1)]
	Aided by the TFRP+SFRP algorithm, the inference time of MetroNext increases by 69ms, and this extra computational overhead is caused by the subnetwork needing to process the left and right splitting proposals, As a result, the model's inference speed is extended to 446ms/frame. Nevertheless, it still maintains a relatively higher image processing speed. 
	\item [2)]
	After implementing the full-mode FRP algorithm, MetroNext's inference time rises to 164ms, and the image processing efficiency of the entire model becomes 541ms/frame. The additional processing procedures, including the RPN for feature analysis of pedestrian proposals and the subnetwork's estimation of pedestrian confidence for these splitting proposals, constrain the model's inference speed. However, this computational cost is acceptable considering the significant enhancement in the model's detection performance. This also suggests that it's suitable to be deployed on high-performing hardware platforms to handle accuracy-prioritized pedestrian detection tasks.
\end{itemize}

It should be mentioned that the Jetson Nano, as an entry-level embedded AI hardware platform, has only 472 GFLOPS of computational power, which limits its capability to speed up the network's inference. When the FRP algorithm is deployed on an advanced edge-computing device, better image processing efficiency will be obtained.

\begin{table}[!h]
	\centering
	\caption{The inference speed of MetroNext with the FRP algorithm on Jetson Nano.}\label{Tbl5}
	\setlength{\tabcolsep}{0.2em}
	\begin{tabular}{lllll}
		\hline
		\multicolumn{4}{l}{\textbf{Models}}             & $\rm{Speed_{CPU}}$(ms/frame) \\ \hline
		\multirow{4}{*}{MetroNext} & TFRP & CFRP & SFRP &          \\
		 &  \ding{55}    &  \ding{55}    &  \ding{55}     & 377      \\
		 &  \ding{51}    &  \ding{55}    &  \ding{51}  & +69      \\
		 &  \ding{51}    &  \ding{51}    &  \ding{51}      & +164     \\ \hline
	\end{tabular}
\end{table}

\section{Conclusion}\label{sec5}
This paper has presented a novel FRP algorithm aimed at effectively suppressing pedestrian false positives generated by the two-stage CNN-based pedestrian detection paradigm under different stages. Three complementary algorithms have been proposed: the TFRP algorithm for the training stage, and the CFRP and SFRP algorithms for the testing stage, which are strategically integrated with the MetroNext to construct two new pedestrian detection networks with balanced prediction performance and strong detection accuracy, ensuring versatility in various applications. Extensive experiments conducted on benchmark datasets and the SY-Metro dataset demonstrate significant advantages of these two new networks for solving pedestrian detection. Deployment experiments on the Jetson Nano further reveal the models' unique advantages: one optimized for speed-critical applications, the other for accuracy-demanding tasks. In short, all experimental results support the feasibility of the FRP algorithm in removing false positives during pedestrian detection and can be used to address pedestrian detection tasks in edge-computing environments.
\newline

\noindent\textbf{Author Contributions} Qiang Guo: Writing-Original draft preparation, Carrying out the experiments, Methodology, Data curation. Rubo Zhang: Supervision, Reviewing and Editing. Bingbing Zhang and Junjie Liu: Contributing to experiments.
\newline

\noindent\textbf{Funding} No funding was received to carry out this study.
\newline

\noindent\textbf{Data availability and access} The benchmark datasets used in this paper come from
these papers \cite{Ref43,Ref44,Ref45}. The SY-Metro dataset are not available due to commercial restrictions.

\section*{Declarations}

\noindent\textbf{Competing Interests} The authors have no competing interests to disclose in any material discussed in this article.
\newline

\noindent\textbf{Ethical and informed consent for data used} Not applicable.
\newline

\bibliographystyle{splncs03}
\bibliography{example}

\end{document}